\documentclass[11pt,a4paper]{article}
\usepackage[dvipsnames,table]{xcolor}
\usepackage[hyperref]{emnlp-ijcnlp-2019}
\usepackage{times}
\usepackage{latexsym}
\usepackage{amsmath}
\usepackage{amssymb}
\usepackage{booktabs}
\usepackage{centernot}
\usepackage{subcaption}
\usepackage{multirow}
\usepackage{url}
\usepackage{ifthen}
\usepackage{xspace}
\usepackage{graphicx}
\usepackage{appendix}

\aclfinalcopy 

\newif\ifcomment
\commenttrue

\newcommand\sz{\ensuremath{\mathcal{z}}}

\newcommand\sD{\ensuremath{\mathcal{D}}}

\newcommand\bt{\ensuremath{\mathbf{t}}}

\newcommand\p[1]{\ensuremath{\left( #1 \right)}} 
\newcommand\pb[1]{\ensuremath{\left[ #1 \right]}} 

\newcommand\R{\ensuremath{\mathbb{R}}} 
\newcommand\eqdef{\ensuremath{\stackrel{\rm def}{=}}} 
\newcommand\refeqn[1]{(\ref{eqn:#1})}

\newcommand\refsec[1]{Section~\ref{sec:#1}}

\newcommand\reffig[1]{Figure~\ref{fig:#1}}

\newcommand\reftab[1]{Table~\ref{tab:#1}}
\newcommand\refapp[1]{Appendix~\ref{sec:#1}}

\ifthenelse{\isundefined{\definition}}{\newtheorem{definition}{Definition}}{}
\ifthenelse{\isundefined{\assumption}}{\newtheorem{assumption}{Assumption}}{}
\ifthenelse{\isundefined{\hypothesis}}{\newtheorem{hypothesis}{Hypothesis}}{}
\ifthenelse{\isundefined{\proposition}}{\newtheorem{proposition}{Proposition}}{}
\ifthenelse{\isundefined{\theorem}}{\newtheorem{theorem}{Theorem}}{}
\ifthenelse{\isundefined{\lemma}}{\newtheorem{lemma}{Lemma}}{}
\ifthenelse{\isundefined{\corollary}}{\newtheorem{corollary}{Corollary}}{}
\ifthenelse{\isundefined{\alg}}{\newtheorem{alg}{Algorithm}}{}
\ifthenelse{\isundefined{\example}}{\newtheorem{example}{Example}}{}
\newcommand{\E}{\ensuremath{\mathbb{E}}} 

\newcommand\nl[1]{``\textit{#1}''}
\newcommand\eg{e.g.,\xspace}

\renewcommand\sup{biased\xspace}
\newcommand\deb{debiased\xspace}
\newcommand\ours{DRiFt\xspace}
\newcommand\relu{ReLU\xspace}
\newcommand\hypo{\textsc{Hypo}\xspace}
\newcommand\hand{\textsc{Hand}\xspace}
\newcommand\cbow{CBOW\xspace}
\newcommand\rmv{\textsc{Rm}\xspace}
\newcommand\ent{$E$\xspace}
\newcommand\neu{$N$\xspace}
\newcommand\con{$C$\xspace}
\newcommand{\nent}{$\neg E$\xspace}
\newcommand{\stress}{\textsc{Stress}\xspace}

\DeclareMathOperator*{\argmin}{arg\,min}
\usepackage{breqn}

\newcolumntype{L}[1]{>{\raggedright\let\newline\\\arraybackslash\hspace{0pt}}m{#1}}
\newcolumntype{C}[1]{>{\centering\let\newline\\\arraybackslash\hspace{0pt}}m{#1}}
\newcolumntype{R}[1]{>{\raggedleft\let\newline\\\arraybackslash\hspace{0pt}}m{#1}}
\renewcommand{\bt}[2]{
	\multicolumn{1}{r}{\cellcolor{cyan!#1}{#2}}
}
\newcommand{\ws}[2]{
	\multicolumn{1}{r}{\cellcolor{red!#1}{#2}}
}

\ifcomment
\newcommand\hh[1]{\textcolor{blue}{[HH: #1]}}
\renewcommand\sz[1]{\textcolor{orange}{[SZ: #1]}}
\else
\newcommand\hh[1]{}
\renewcommand\sz[1]{}
\fi

\title{Unlearn Dataset Bias in Natural Language Inference by Fitting the Residual}

\author{He He$^{1,2}$ \and Sheng Zha$^{1}$ \and Haohan Wang$^{3}$ \\
    $^1$Amazon Web Services, $^2$New York University, $^3$Carnegie Mellon University \\
    {\tt \{hehea,zhasheng\}@amazon.com, haohanw@cs.cmu.edu}}

\date{}

\begin{document}
\maketitle
\begin{abstract}
    Statistical natural language inference (NLI) models are susceptible to learning \emph{dataset bias}: superficial cues that happen to associate with the label on a particular dataset, but are not useful in general,
    \eg negation words indicate contradiction.
    As exposed by several recent challenge datasets,
    these models perform poorly when such association is absent,
    \eg predicting that \nl{I love dogs.} contradicts \nl{I don't love cats.}.
    Our goal is to design learning algorithms that guard against \emph{known} dataset bias.
    We formalize the concept of dataset bias under the framework of distribution shift and present a simple debiasing algorithm based on residual fitting, which we call \ours.
    We first learn a biased model that only uses features that are known to relate to dataset bias.
    Then, we train a debiased model that fits to the residual of the biased model,
    focusing on examples that cannot be predicted well by biased features only.
    We use \ours to train three high-performing NLI models on two benchmark datasets, SNLI and MNLI.
    Our debiased models achieve significant gains over baseline models on two challenge test sets,
    while maintaining reasonable performance on the original test sets.\footnote{
    Code is available at \url{https://github.com/hhexiy/debiased}.}
\end{abstract}

\section{Introduction}
\label{sec:intro}
Machine learning models have surpassed human-performance on multiple language understanding benchmarks.
However, transferring the success to real-world applications has been much slower
due to the brittleness of these systems.
For example, \newcite{mccoy2019hans} show that models blindly predict the entailment relation for two sentences with high word overlap even if they have very different meanings, \eg \nl{The man hit a dog} and \nl{The dog hit a man}.
\newcite{jia2017adversarial} show that reading comprehension models are easily distracted by irrelevant sentences containing key phrases from the question.
Similar failures have also been observed on paraphrase identification~\cite{zhang2019paws} and story cloze test~\cite{schwartz2017roc}.

\begin{figure}[t]
    \centering
\includegraphics[width=0.9\columnwidth]{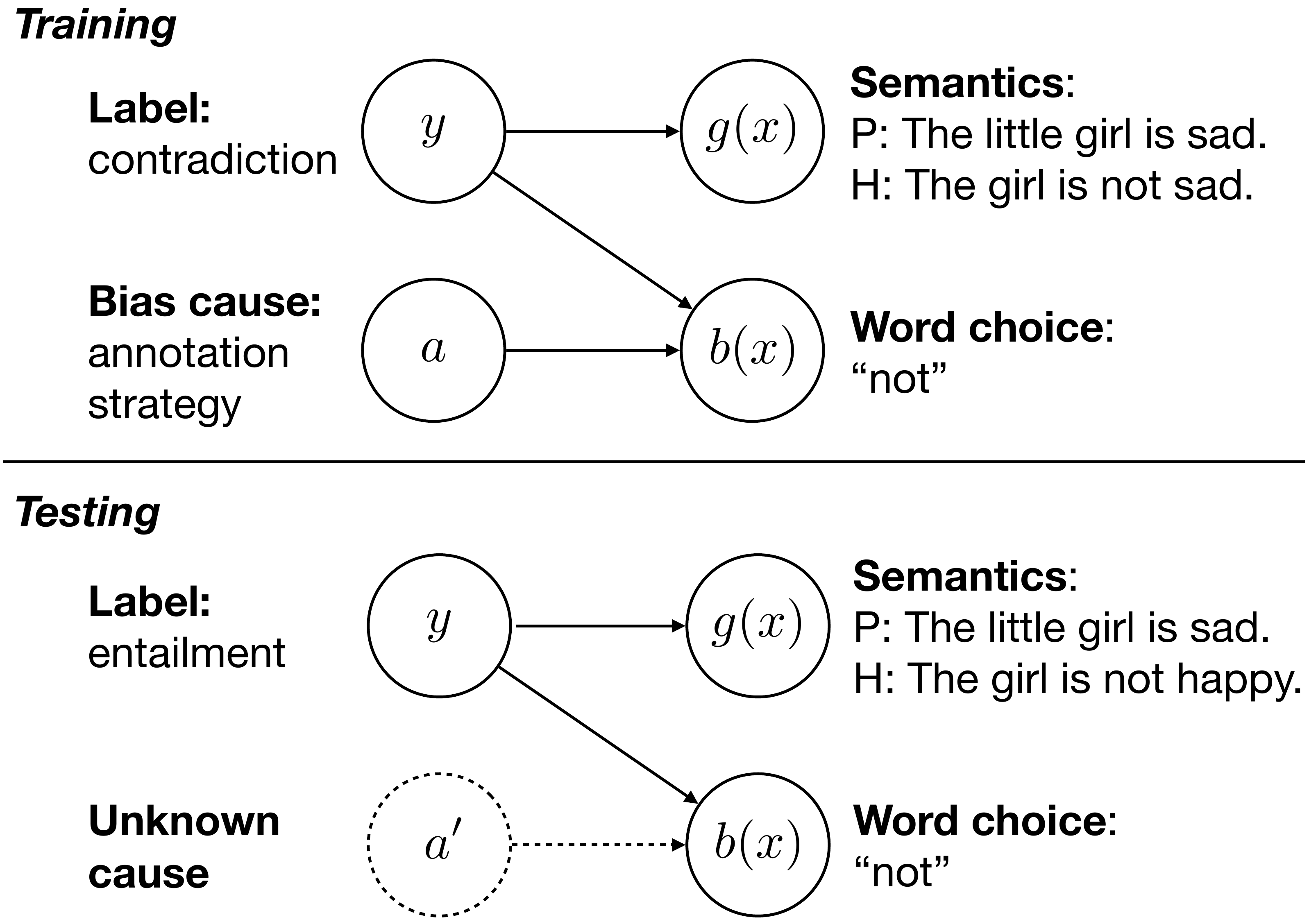}
\caption{An example of dataset bias in NLI.
On the training data, the biased feature (\nl{not}) is affected by crowd workers' strategy of negating the premise to create a contradicting pair.
However, at test time the word choice is affected by \emph{unknown} sources,
thus \nl{not} may not be associated with the label ``contradiction''.
A model relying on the negation word to predict ``contradiction'' would fail on the shown test example.
}
\label{fig:dist-shift}
\end{figure}

A common problem behind these failures is distribution shift. 
Our training data is often not a representative sample of real-world data due to their different data-generating processes,
thus models are susceptible to learning simple cues (\eg lexical overlap) that work well on the majority of training examples but fail on more challenging test examples.
Consider generating a contradicting pair of sentences for natural language inference (NLI) in \reffig{dist-shift}.
Crowd workers tend to mechanically negate the premise sentence to save time,
introducing an association between negation words (\eg \nl{not}) and the contradiction label.
However, at test time, such association may not exist as data is now generated by end users.
Thus, a model that heavily relies on the biased feature \nl{not} would fail.
In this paper, we formalize \emph{dataset bias}~\cite{torralba2011unbiased} under the label shift assumption:
the conditional distribution of the label given biased features changes at test time.
Our goal is to design learning algorithms that are robust to dataset bias with a focus on NLI,
i.e. predicting whether the \emph{premise} sentence entails the \emph{hypothesis} sentence.

Typical debiasing approaches aim to remove biased features (\eg gender and image texture) in the
learned representation~\cite{wang2019balanced,wang2019learning}.
However, biased features in textual data often conflate useful semantic information and superficial cues,
thus completely removing them might significantly hurt prediction performance.
Even when we are confident that the bias is irrelevant to prediction (\eg gender),
\citet{gonen2019lipstick} show that existing bias removal methods are insufficient.

Instead of debiasing the data representation, our method (along with the concurrent work of \citet{clark2019ensemble}) accounts for label shift given biased features
by focusing on ``hard'' examples that cannot be predicted well using only biased features.
We train a model in two steps.
First, we train a \emph{\sup model} using insufficient features such as overlapping words between the premise and the hypothesis.
Next, we train a \emph{\deb model} by fitting to the residuals of the \sup model.
This step ``unlearns'' the bias by taking additional negative gradient updates on
examples with low loss under the biased model (\refsec{analysis}).\footnote{
    Note that dataset bias is flagged by good performance despite insufficient input, \eg a high-accuracy hypothesis-only classifier~\cite{gururangan2018annotation}.
}
At test time, only the \deb model is used for prediction.
We call this learning algorithm \ours (\textbf{D}ebias by \textbf{R}es\textbf{i}dual \textbf{F}i\textbf{t}ting).

We use \ours to train three high-performing NLI models on two benchmark datasets, SNLI~\cite{bowman2015large} and MNLI~\cite{williams2017broad}.
Compared to baseline models trained by maximum likelihood estimation,
our debiased models improve performance
on several challenge datasets
with only slight degradation on the original test sets.

\section{Problem Statement}
\label{sec:problem}
\paragraph{Dataset bias.}
Let $x \in \mathcal{X}$ be the input and $y \in \mathcal{Y}$ be the label we want to predict.
Given training examples $(x, y)$ drawn from a distribution $P$,
we define \emph{dataset bias} as (partial) representation of $x$ that exhibits label
shift~\cite{lipton2018detecting, scholkopf2012causal} on the test distribution $Q$.
Formally, assume that $x$ can be represented by two components $b(x)$ and $g(x)$ conditionally independent given $y$.
We have
\begin{align}
p(x, y) &= p(b(x), g(x), y) \\
        &= p(g(x) \mid y) p(y \mid b(x)) p(b(x)) .
\end{align}
Let $g(x)$ be the true effect of $y$ such that their relationship does not change normally, i.e.
$p(g(x) \mid y) = q(g(x) \mid y)$.
Let $b(x)$ be \emph{\sup features} that happen to be predictive of $y$ on $P$.
For example, in \reffig{dist-shift},
$g(x)$ represents semantics of the premis and hypothesis sentences,
whereas $b(x)$ represents specific word choices affected by varying sources.
In the training data, the word \nl{not} has a strong association with ``contradiction''
due to crowd workers' writing strategies.
Consequently, a model learned on the training data distribution $P$ would degrade when such association no longer exists.
Formally, both training and testing examples may exhibit biased features: $p(b(x)) = q(b(x))$,
but dependence between these features and the label can change: $p(y \mid b(x)) \neq q(y \mid b(x))$.

In a typical supervised learning setting with dataset bias,
we do not observe examples from $Q$ thus $b(x)$ is unknown.
Without additional information, achieving good performance on $Q$ is impossible.
Fortunately, oftentimes we do have domain-specific knowledge on what $b(x)$ might be,
\eg the word overlapping heuristic in NLI.
Therefore, our goal is to correct the model trained on $P$ to perform well on $Q$ given \emph{known} dataset bias.

\paragraph{Bias in NLI data.}
Dataset bias in SNLI~\cite{bowman2015large} and MNLI~\cite{williams2017broad} are largely due to the crowdsourcing process.
Both are created by asking crowd workers to write three sentences (hypotheses) that are entailed by, neutral with, or contradict a given sentence drawn from a corpus (the premise). 
\citet{gururangan2018annotation,poliak2018hypothesis} show that certain words in the hypothesis have high pointwise mutual information with class labels regardless of the premise,
which could be artifacts of specific annotation strategies.
For example, one can create a neutral sentence by adding a cause (\nl{because}) to the premise
and create a contradicting sentence by negating (\nl{no}, \nl{never}) the premise.
As a result, the majority of training examples can be solved without much reasoning about sentence meanings.
Subsequently, \citet{mccoy2019hans} report that models rely on high word overlap to predict entailment;
\citet{glockner2018breaking,naik2018stress} demonstrate that models struggle at even lexical-level inference involving antonyms, hypernyms, etc.

A natural question to ask then is whether there exist better data collection procedures that guard against these biases.
We argue that this is not easy
because in practice, we almost always have different data-generating processes during training (generated from selected corpora and annotators) and test (generated by end users).
Then, can we remove biased features from training examples?
This is also infeasible because sometimes they contain the necessary information for prediction,
\eg removing words may destroy the sentence meaning.
It is not the features that are biased but their relation with the label.
Next, we describe our approach to mitigating this biased relation.

\section{Approach}
\label{sec:approach}
\subsection{Overview}
The key idea of our approach is to first detect biased examples given prior knowledge on potential dataset bias,
then focus on learning from unbiased, hard examples.
We describe the two steps in details below.

\paragraph{Detect biased examples.}
How do we know if an example exhibits biased features?
Although we cannot directly measure label shift without accessing the test data,
we know that NLI models are unlikely to work well given insufficient features.
When it does work well given only partial semantics of the input,
the good performance is likely due to dataset bias.
For example,
\citet{gururangan2018annotation} exposes annotation artifacts by showing that hypothesis-only models
have unexpected high accuracy.
Similarly, we train a \emph{\sup classifier} using insufficient features $I(x)$, \eg the hypothesis sentence.
We assume that examples predicted well by the biased classifier exhibit dataset bias,
i.e. $p(y \mid I(x))$ is high but $q(y \mid I(x))$ is low.

Importantly, while $I(x)$ approximates $b(x)$ given our prior knowledge,
it does not necessarily capture all dataset bias, which depends on the unknown test distribution.
In addition, $I(x)$ may include useful information.
For example, although bag-of-words (BOW) features are insufficient to represent precise sentence meaning,
it encodes a distribution of possible meanings.
Thus good performance of a BOW classifier is not fully due to fitting dataset bias.
In practice, as we will see in the experiments (\refsec{hans}),
good choices of $I(x)$ capture biased features precisely,
resulting in significant performance drop of the biased classifier on $Q$.

\paragraph{Learn residuals of the \sup classifier.}
Our intuition is that the \deb classifier should capture information beyond those
contained in the \sup classifier.
If the \sup classifier already has a small loss on an example, then there is not much to learn beyond the biased features;
otherwise, the \deb classifier should correct predictions of the biased classifier.

We implement the idea through a residual fitting procedure (\ours).
Let $f_s\colon \mathcal{X} \rightarrow \mathbb{R}$ and $f_d\colon \mathcal{X} \rightarrow \mathbb{R}$ be the \sup and the \deb classifiers,
and let $L$ be the loss function.
First, we learn $f_s$ with insufficient features $I(x)$ as the input:
\begin{align}
\theta^* = \argmin_\theta \E_P\pb{L(f_s(I(x); \theta), y)} .
\end{align}
Let $f^*(x)$ be the optimal predictor that minimizes the empirical risk on $P$.
We define
\begin{align}
f^*(x) \eqdef f_s(I(x); \theta^*) + f_d(x; \phi^*) .
\end{align}
Thus $f_d$ fits the residual of $f_s$ with respect to the target $f^*$.
To estimate parameters $\phi$ of $f_d$, we fix parameters of $f_s$ and minimize the loss:
\begin{align}
    \min_\phi \E_P\pb{L(f_s(I(x); \theta^*) + f_d(x; \phi), y)} \label{eqn:loss}.
\end{align}
At test time, we only use the \deb classifier $f_d$.

Consider the typical empirical risk minimization approach
that estimates $\phi$ by minimizing $\E_P\pb{L(f_d(x; \phi), y)}$.
It is susceptible to relying on biased features when they predict well on the majority examples.
In contrast, \ours first learns $f_s$ which is intended to fit potential bias in the data.
It then learns $f_d$ 
that compensates $f_s$ without fitting to the bias already captured by it.

Next, we analyze the behavior of \ours using the cross-entropy loss function,
which is typically used for classification problems.

\subsection{Analysis with the Cross-Entropy Loss}
\label{sec:analysis}
In this section, we show that \ours adjusts the gradient on each example depending on how well it is predicted by the pretrained biased classifier.

Given the cross-entropy loss, our goal is to maximize the expected conditional log-likelihood of the data,
$\E_P\pb{\log p(y\mid x)}$.
A classifier outputs a vector of scores for each of the $K$ classes, $f(x) = (f^1(x), \ldots, f^K(x)) \in \R^K$,
which are then mapped to a probability distribution $p(y\mid x)$ by the softmax function.
Given classifiers $f_s$ and $f_d$, we have three choices of parametrization of the conditional probability $p(y \mid x)$:
\begin{align}
    p_s(y \mid I(x)) &\propto {\exp{\p{f_s^y(I(x); \theta)}}} \label{eqn:ps}\\
    p_d(y \mid x) &\propto {\exp{\p{f_d^y(x; \phi)}}} \label{eqn:pd}\\
    p_a(y \mid x) &\propto {\exp{\p{f_s^y(I(x); \theta) + f_d^y(x; \phi)}}} \notag \\
                  &\propto {p_s(y \mid I(x)) p_d(y \mid x)} . \label{eqn:pa2}
\end{align}
To learn the classifier $f_d$, standard maximum likelihood estimation (MLE) uses $p_d(y \mid x)$,
whereas \ours uses $p_a(y \mid x)$ given pretrained $f_s$ with fixed parameters.

Let us first compare the two learning objectives.
Denote $p_s(y \mid I(x); \theta^*)$ by $p_s^*(y \mid I(x))$.
\ours maximizes
\begin{align}
    J_{\text{D}}(\phi) &= \sum_{(x,y)\sim \sD} \log p_a(y \mid x; \theta^*,\phi) \\
                         &= C + \sum_{(x,y)\sim \sD} [ \log p_d(y\mid x; \phi) - \notag \\
                         &\phantom{={}} \log\sum_{k=1}^K p_s^*(k\mid I(x))p_d(k\mid x; \phi) ] \label{eqn:JD} \;,
\end{align}
where $\sD$ denotes the training set and $C=\sum_{(x,y)\sim \sD} \log p_s^*(k \mid I(x))$ is a constant.
Compare \refeqn{JD} with the MLE objective:
\begin{align}
    J_{\text{MLE}}(\phi) = \sum_{(x,y)\sim \sD} \log p_d(y \mid x; \phi) \;.
\end{align}
We see that $J_{\text{D}}(\phi)$ has an additional regularizer for each example $x$:
\begin{align}
    R(x) \eqdef -\log\sum_{k=1}^K p_s^*(k\mid I(x))p_d(k\mid x) \;.
\end{align}
Geometrically, it encourages output from the debiased classifier, $p_d$,
to have minimal projection on $p_s$ predicted by the biased classifier.

Next, let's look at the effect of this regularizer through its gradient.
Let $Z(x)$ be the normalizer $\sum_{k} p_s^*(k\mid I(x))p_d(k \mid x)$. Then, we have
\begin{align}
    \nabla_\phi R(x) &= -\frac{\sum_{k} p_s^*(k\mid I(x))\nabla_\phi p_d(k \mid x)}{\sum_{k} p_s^*(k\mid I(x))p_d(k \mid x)} \notag \\
    &= -\sum_{k} p_a(k \mid x)\nabla_\phi \log p_d(k \mid x) \notag ,
\end{align}
which is derived by writing $\nabla_\phi p_d$ as $p_d\nabla_\phi \log p_d$.
Taking a negative step in the direction of $\nabla_\phi \log p_d(k \mid x)$
corresponds to down-weighting the probability $p_d(k \mid x)$.
Intuitively, the model tries to reweight the output distribution by the gradient weights $p_a(k \mid x)$.
Note that
\begin{align}
    p_a(k \mid x) \propto p_s^*(k \mid I(x))p_d(k \mid x) \;.
\end{align}
For an example $(x, y)$,
large values of $p_s^*(y \mid I(x))$ indicate that $I(x)$ is likely to contain biased features.
If $p_d(y \mid x)$ is also large,
the model is probably picking up the bias since
$p_d$ has access to complete information in $x$ including the biased features,
in which case a relatively large negative step is taken to correct it.
In the extreme case where the biased classifier makes perfect prediction,
we have $p_s^*(y \mid I(x)) \to 1$
thus $\nabla_\phi R(x) \to - \nabla_\phi\log p_d(y \mid x)$,
canceling the MLE gradient $\nabla_\phi\log p_d(y \mid x)$.
As a result, the gradient on this example is zero, and there is nothing to be learned.
At the other end where $I(x)$ does not provide any useful information,
the biased classifier outputs a uniform distribution $p_s^*(y \mid I(x)) = 1 / K$,
thus $p_a(y\mid x) = p_d(y\mid x)$
and the gradient on this example is reduced to the MLE gradient.



\section{Experiments}
\label{sec:experiments}
We first evaluate our method using synthetic bias to show its effectiveness under different amount of dataset bias.
We then test on two challenge datasets using different biased classifiers.
We show that \ours consistently outperforms MLE on the challenge datasets given different NLI models,
especially when the insufficient features capture dataset bias exploited by the challenge data.

\subsection{Training Data}
We evaluate \ours on two benchmarking NLI datasets:
SNLI~\cite{bowman2015large} and MNLI~\cite{williams2017broad}.
Each pair of premise and hypothesis sentences has a label from one of
``entailment'', ``contradiction'', or ``neutral''.
Sentences from SNLI are derived from image captions,
whereas MNLI covers a broader range of styles and topics.
Statistics of the two datasets are shown in \reftab{datasets-stats}.
All MNLI results are on the matched development set.\footnote{
    MNLI has two development sets, one from the same source as the training data (matched)
    and one from different sources (mismatched).
    We trained two sets of models using their corresponding development sets
    for model selection
    and obtained similar results.
    Thus we focus on the ``matched'' results.
}

\begin{table}
\centering
\footnotesize{
\begin{tabular}{lrrr}
\toprule
Dataset & Train & Dev & Test \\
\midrule
SNLI & 549,367 & 9842 & 9842 \\
MNLI & 392,702 & 9815 & - \\
\bottomrule
\end{tabular}
}
\caption{Statistics of training datasets.
The test sets of MNLI are hosted through Kaggle competitions.
}
\label{tab:datasets-stats}
\end{table}

\subsection{Models and Training Details}
\label{sec:details}
\ours is a general learning algorithm that works with any biased/debiased models.
Below we describe the three key components of our approaches:
the learning algorithm, the biased model with its insufficient features, and the debiased model.

\paragraph{Learning algorithms.}
We compare \ours with MLE, as well as
a simpler variant of \ours:
instead of the residual fitting, we remove the examples predicted correctly by the \sup classifier and train on the rest.
We call this baseline \rmv,
which is also conceived by \citet{gururangan2018annotation}.
MLE only trains the debiased model.
Both \ours and \rmv rely on an additional \sup model that captures potential dataset bias.

\paragraph{Biased models.} We consider three insufficient representations that exploit various NLI dataset biases reported in prior work.
\paragraph{\hypo} is a finetuned BERT classifier that uses only the hypothesis sentence.
\paragraph{CBOW} is a continuous bag-of-words classifier.
Similar to \citet{hou2016natural}, we represent both the premise and the hypothesis as the respective sums of their word embeddings.
We then concatenate the premise and the hypothesis embeddings,
their difference, and their element-wise product.
The final representation is passed through a one-layer fully connected network with \relu activation.
\paragraph{\hand} is a classifier using handcrafted features based on error analysis in \citet{naik2018stress}.
Specifically, we include tokens in the hypothesis that are also in the premise,
tokens unique to the hypothesis,
Jaccard similarity between the two sentences,
whether negation words (\nl{not} and \nl{n't}) are included,
and length difference computed by 
$\frac{|L_p - L_h|}{L_p + L_h}$ where
$L_p$ and $L_h$ are numbers of tokens in the premise and the hypothesis.
We represent the overlapping and the non-overlapping tokens as the respective sums of their word embeddings.
The embeddings are then concatenated with the dense features
and passed through a one-layer fully connected network with \relu activation.

\paragraph{Debiased models.} We choose three high-performing models of different capability.
\paragraph{DA} is the Decomposable Attention model introduced by \citet{parikh2016decomposable},
which relies on the interaction between words in the premise and the hypothesis. It does not use any word order information.
We used the variant without intra-sentence attention.\footnote{
    We removed the projection layers of the word embeddings as it speeds up training without hurting performance in our experiments.
}
\paragraph{ESIM} is the Enhanced Sequential Inference Model~\cite{chen2017enhanced}.
It first encodes the premise and the hypothesis by a bidirectional LSTM,
aligns the contextual word embeddings similar to \citet{parikh2016decomposable},
and uses another ``inference'' bidirectional LSTM to aggregate information.
Thus it has access to the non-local context.
\paragraph{BERT} is the Bidirectional Encoder Representations from Transformers~\cite{devlin2019BERT} that recently improved performance on MNLI significantly.
It uses contextual embeddings pretrained from large corpora.

\paragraph{Hyperparameters.}
For non-BERT models, word embeddings are initialized with the \texttt{840B.300d} pretrained GloVe~\cite{pennington2014glove} word vectors and finetuned during training.
For DA and ESIM, hyperparameters of the model architecture are the same as those reported in the original papers.
We finetune all BERT models from the pretrained \texttt{BERT-base-uncased} model.\footnote{
    \url{http://gluon-nlp.mxnet.io/model_zoo/bert/index.html}}
We train all models using the Adam~\cite{kingma2014adam} optimizer with $\beta_1=0.9$, $\beta_2=0.999$, L2 weight decay of 0.01, learning rate warmup for the first $10\%$ of updates and linear decay afterwards.
We use a dropout rate of 0.1 for all models except ESIM, which has a dropout rate of 0.5.
BERT and non-BERT models are trained with a learning rate of 2e-5 and 1e-4, respectively.
For MLE, we train BERT for 4 epochs and the rest for 30 epochs.
When training the \deb model in \ours, we find that the models converge slowly
thus we train BERT for 8 epochs and the rest for 80 epochs.

\begin{figure*}[t]
    \includegraphics[width=\textwidth]{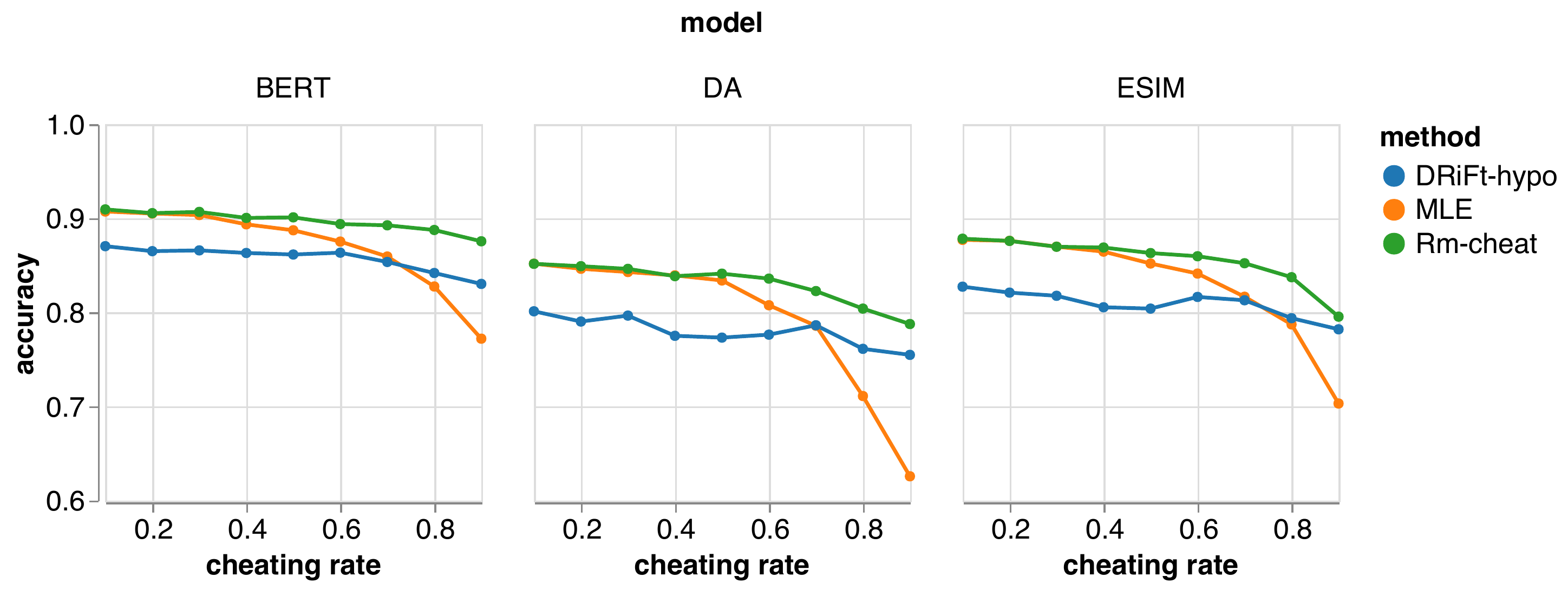}
    \caption{Accuracy on SNLI test set augmented with cheating features,
        which leak the groundtruth labels on training data
        but not on test data.
        Models trained by MLE degrade significantly when a majority of examples are cheatable,
        whereas \deb models trained by \ours maintain similar accuracies across different cheating rates.
    }
    \label{fig:cheat}
\end{figure*}

\subsection{In-Distribution Performance}
We first evaluate the models' in-distribution performance where
they are trained and evaluated on splits from the same dataset.
Results of the \sup models are reported in \reftab{sup}.
All exceeds the majority-class baseline by a large margin,
indicating that a majority of examples can be solved by superficial cues.

Results of the \deb models are reported in \reftab{in-dist}.
Baseline results from our implementations are comparable to prior reported performance (row ``MLE'').
Debiased models trained by \ours show some degradation on in-distribution data,
especially for the less powerful DA and ESIM models.
The accuracy drop is expected due to two reasons.
First, \ours assumes distribution shift thus does not optimize performance on the training distribution $P$.
Second, the effective training data size is reduced by negative gradients on potentially biased examples;
this effect is exaggerated by \rmv, which shows significant in-distribution degradation.
Similar trade-off between in-distribution accuracy and robustness on out-of-distribution data has also been observed in adversarial training~\cite{zhang2019theoretically,tsipras2019robustness}.

\begin{table}[th]
    \centering
\footnotesize{
\begin{tabular}{lrrrr}
    \toprule
    Dataset & majority & \hypo & \cbow & \hand \\
    \midrule
    SNLI & 34.2 & 61.8 & 81.2 & 76.7 \\
    MNLI & 35.4 & 52.5 & 66.1 & 65.4 \\
    \bottomrule
\end{tabular}
}
\caption{Accuracy of \sup classifiers on SNLI test set and MNLI development set.
All exceeds the majority-class baseline by a large margin,
signaling dataset bias.
}
\label{tab:sup}
\end{table}

\begin{table}[t]
\centering
\footnotesize{
\setlength{\tabcolsep}{3pt}
\begin{tabular}{lrrrrrr}
\toprule
    		& \multicolumn{3}{c}{SNLI} & \multicolumn{3}{c}{MNLI} \\
			  \cmidrule(lr){2-4} \cmidrule(lr){5-7}
    		& BERT            & DA              & ESIM             & BERT           & DA              & ESIM   \\
\midrule
MLE & \bt{0.0}{90.8} & \bt{0.0}{85.3} & \bt{0.0}{88.0} & \bt{0.0}{84.5} & \bt{0.0}{72.2} & \bt{0.0}{78.1} \\
\midrule
\ours-\hypo & \ws{1.0}{89.8} & \ws{1.3}{83.9} & \ws{1.6}{86.3} & \ws{0.1}{84.3} & \ws{3.6}{68.6} & \ws{3.1}{75.0} \\
\ours-\cbow & \ws{6.1}{84.7} & \ws{22.7}{62.6} & \ws{25.7}{62.3} & \ws{2.3}{82.1} & \ws{15.9}{56.3} & \ws{9.3}{68.8} \\
\ours-\hand & \ws{4.3}{86.5} & \ws{10.2}{75.0} & \ws{8.8}{79.2} & \ws{2.7}{81.7} & \ws{13.3}{58.8} & \ws{9.2}{68.9} \\
\midrule
\rmv-\hypo & \ws{19.6}{71.2} & \ws{18.3}{67.0} & \ws{17.7}{70.3} & \ws{19.0}{65.5} & \ws{14.7}{57.5} & \ws{15.1}{63.0} \\
\rmv-\cbow & \ws{55.0}{35.8} & \ws{58.2}{27.1} & \ws{65.8}{22.2} & \ws{29.6}{54.9} & \ws{45.4}{26.8} & \ws{51.0}{27.1} \\
\rmv-\hand & \ws{44.5}{46.3} & \ws{48.1}{37.2} & \ws{49.9}{38.1} & \ws{32.7}{51.7} & \ws{37.6}{34.6} & \ws{40.7}{37.4} \\
\bottomrule
\end{tabular}
}
\caption{Accuracy of models trained by MLE, \ours, and
\rmv with different \sup models.
Training and test examples are from the same dataset.
Intensity of the \colorbox{red!50}{red} highlights corresponds to \emph{absolute} drop in accuracy with respect to the MLE baseline.
\rmv significantly hurts in-distribution performance.
\ours maintains reasonable performance.
}
\label{tab:in-dist}
\end{table}

\subsection{Synthetic Bias}
In this section, we evaluate our model under controlled, synthetic dataset bias on SNLI.
Recall our definition of dataset bias:
the conditional distribution of the label $y$ given biased features are different on training and test sets.
Therefore, we inject bias into each example by adding a \emph{cheating feature} that encodes its label.
On training and development examples, the cheating feature encodes the ground truth label with probability $p_{\text{cheat}}$ (the cheating rate),
and a random label otherwise.
On test examples, the cheating feature always encodes a random label.
Thus a model relying on the cheating feature would perform poorly on the test set.

Specifically, we prepend the hypothesis with a string ``\{\texttt{label}\} \textit{and}'' where $\texttt{label} \in \{\text{entailment, contradiction, neutral}\}$.
To simulate the fact that we often cannot pinpoint biased features until the model fails on some test examples,
we choose \hypo as our biased classifier.
That is to say, we have a rough idea that the bias might be in the hypothesis but do not know what it is exactly.

We train all three base models (DA, ESIM, and BERT) using MLE
and \ours, respectively.
Our results are shown in \reffig{cheat}.
All MLE models are reasonably robust to a mild amount of bias.
However, when a majority ($p_{\text{cheat}} > 0.6$) of training examples contains the bias, their accuracy decreases significantly:
about $20\%$ drop at $p_{\text{cheat}} = 0.9$ compared to the baseline accuracy when no cheating features are injected.
BERT is slightly more robust than DA and ESIM, possibly due to the regularization effect of pretrained embeddings.
In contrast, our debiased models (\ours-\hypo) maintain similar accuracies with increasing cheating rates and have a maximum accuracy drop of about $5\%$.

Two questions remain, though:
(1) Why does the accuracy of debiased models still drop a bit at high cheating rates?
(2) Why is the baseline accuracy of \ours lower than MLE?
We answer these questions by analyzing the upper bound performance of our method below.

\paragraph{Best-case scenario.}
In the ideal case, we know precisely what the bias is.
Consider a biased classifier that only uses the cheating feature as its input.
It predicts biased examples perfectly,
i.e. $p_s(y\mid b(x)) = 1$ and $p_s(k\mid b(x)) = 0 \;\forall k \neq y$,
and predicts the rest unbiased examples uniformly at random.
Based on our discussion at the end of \refsec{analysis},
the biased examples have zero gradients 
and unbiased examples have the same gradients as in MLE.
In this case, our method is equivalent to removing biased examples and training a classifier on the rest, i.e. \rmv-cheat.
In \reffig{cheat}, we see that it completely dominates MLE.
The accuracy of \rmv-cheat still drops when $p_{\text{cheat}}$ is large,
because there are fewer ``good'' examples to learn from,
not due to fitting the bias.
Similarly, \ours-\hypo has lower overall accuracy compared to \rmv-cheat,
because \hypo captures additional (unbiased) features
that cannot be fully learned by the \deb model.

\paragraph{Worst-case scenario.}
In the extreme case when $p_{\text{cheat}}=1$, all models' predictions on the test set are random guesses.
For MLE, the biased features are no longer differentiable from the generalizable ones,
thus there is no reason not to use them.
For \ours, since the \sup model achieves perfect prediction on all training examples,
the \deb model receives zero gradient.
Therefore, when strong bias presents on all examples,
we need more information to correct the bias,
\eg collecting additional data or augmenting examples.

\begin{table}[ht]
    \centering
\footnotesize{
\setlength{\tabcolsep}{4pt}
\begin{tabular}{lrrrrrr}
\toprule
\multirow{2}{*}{method} & \multicolumn{2}{c}{lexical} & \multicolumn{2}{c}{subseq} & \multicolumn{2}{c}{const} \\
     \cmidrule(lr){2-3} \cmidrule(lr){4-5} \cmidrule(lr){6-7}
            & \ent            & \nent           & \ent            & \nent           & \ent             & \nent \\
\midrule
\hypo       & \bt{0.0}{52.6} & \bt{0.0}{44.4} & \bt{0.0}{54.5} & \bt{0.0}{44.3} & \bt{0.0}{45.6} & \bt{0.0}{16.7} \\
\cbow       & \bt{0.0}{63.2} & \bt{0.0}{16.0} & \bt{0.0}{66.2} & \bt{0.0}{33.7} & \bt{0.0}{63.2} & \bt{0.0}{38.5} \\
\hand       & \bt{0.0}{66.7} & \bt{0.0}{0.0} & \bt{0.0}{66.7} & \bt{0.0}{0.0} & \bt{0.0}{66.7} & \bt{0.0}{0.0} \\
\midrule
\multicolumn{7}{l}{\textbf{model: BERT}} \\
\midrule
MLE & \bt{0.0}{67.2} & \bt{0.0}{7.8} & \bt{0.0}{66.7} & \bt{0.0}{0.4} & \bt{0.0}{68.1} & \bt{0.0}{11.9} \\
\midrule
\ours-\hypo & \bt{17.5}{84.7} & \bt{72.1}{79.8} & \bt{2.3}{69.0} & \bt{23.3}{23.7} & \bt{4.7}{72.7} & \bt{28.9}{40.8} \\
\ours-\cbow & \bt{13.6}{80.8} & \bt{67.4}{75.2} & \bt{1.8}{68.5} & \bt{29.1}{29.5} & \bt{3.4}{71.5} & \bt{28.4}{40.3} \\
\ours-\hand & \bt{10.3}{77.4} & \bt{63.2}{70.9} & \bt{4.5}{71.2} & \bt{40.8}{41.2} & \bt{7.8}{75.8} & \bt{49.1}{61.0} \\
\midrule
\rmv-\hypo & \bt{0.0}{67.2} & \bt{38.2}{46.0} & \ws{1.5}{65.2} & \bt{36.2}{36.6} & \bt{7.4}{75.5} & \bt{60.3}{72.2} \\
\rmv-\cbow & \ws{61.7}{5.4} & \bt{58.6}{66.4} & \ws{58.2}{8.5} & \bt{63.8}{64.2} & \ws{33.3}{34.8} & \bt{53.4}{65.3} \\
\rmv-\hand & \ws{57.1}{10.0} & \bt{58.3}{66.0} & \ws{62.0}{4.7} & \bt{65.9}{66.3} & \ws{59.0}{9.1} & \bt{55.4}{67.3} \\
\midrule
\multicolumn{7}{l}{\textbf{model: DA}} \\
\midrule
MLE & \bt{0.0}{66.6} & \bt{0.0}{0.5} & \bt{0.0}{66.6} & \bt{0.0}{0.3} & \bt{0.0}{66.5} & \bt{0.0}{0.4} \\
\midrule
\ours-\hypo & \ws{0.3}{66.3} & \bt{1.2}{1.7} & \bt{0.3}{66.9} & \bt{5.2}{5.5} & \ws{0.2}{66.3} & \bt{8.0}{8.4} \\
\ours-\cbow & \ws{1.3}{65.3} & \bt{6.7}{7.2} & \ws{0.5}{66.1} & \bt{9.3}{9.6} & \ws{1.4}{65.1} & \bt{8.7}{9.1} \\
\ours-\hand & \ws{6.1}{60.5} & \bt{26.7}{27.1} & \ws{5.2}{61.4} & \bt{44.6}{44.9} & \ws{10.6}{55.9} & \bt{47.9}{48.3} \\
\midrule
\rmv-\hypo & \ws{1.5}{65.1} & \bt{9.1}{9.6} & \ws{0.4}{66.2} & \bt{14.7}{15.0} & \ws{0.2}{66.2} & \bt{18.4}{18.8} \\
\rmv-\cbow & \ws{66.2}{0.4} & \bt{66.2}{66.6} & \ws{65.4}{1.3} & \bt{66.4}{66.7} & \ws{65.6}{0.8} & \bt{66.1}{66.5} \\
\rmv-\hand & \ws{56.3}{10.3} & \bt{65.3}{65.8} & \ws{57.7}{8.9} & \bt{65.4}{65.7} & \ws{52.6}{13.9} & \bt{64.4}{64.7} \\
\midrule
\multicolumn{7}{l}{\textbf{model: ESIM}} \\
\midrule
MLE & \bt{0.0}{65.8} & \bt{0.0}{3.2} & \bt{0.0}{67.2} & \bt{0.0}{4.6} & \bt{0.0}{65.5} & \bt{0.0}{2.8} \\
\midrule
\ours-\hypo & \ws{1.6}{64.3} & \bt{7.3}{10.5} & \bt{1.1}{68.3} & \bt{11.7}{16.3} & \bt{2.5}{68.1} & \bt{26.5}{29.3} \\
\ours-\cbow & \ws{2.6}{63.2} & \bt{11.3}{14.4} & \ws{0.4}{66.8} & \bt{15.5}{20.1} & \ws{0.6}{64.9} & \bt{19.9}{22.7} \\
\ours-\hand & \ws{4.6}{61.2} & \bt{16.4}{19.6} & \ws{3.5}{63.7} & \bt{34.8}{39.4} & \ws{0.8}{64.8} & \bt{45.5}{48.3} \\
\midrule
\rmv-\hypo & \ws{2.6}{63.3} & \bt{9.6}{12.8} & \ws{3.1}{64.1} & \bt{20.2}{24.8} & \bt{5.8}{71.3} & \bt{43.2}{46.0} \\
\rmv-\cbow & \ws{61.3}{4.5} & \bt{62.5}{65.7} & \ws{61.2}{6.0} & \bt{60.6}{65.2} & \ws{48.6}{16.9} & \bt{61.0}{63.8} \\
\rmv-\hand & \ws{40.0}{25.8} & \bt{57.6}{60.8} & \ws{48.9}{18.3} & \bt{62.7}{67.3} & \ws{52.4}{13.1} & \bt{63.1}{65.9} \\
\bottomrule
\end{tabular}
}
\caption{F1 scores of the entailment (\ent) and non-entailment (\nent) classes on HANS.
    All models are trained on MNLI and
    results are shown on three subsets targeting at different biases:
    lexical overlap (lexical), subsequence overlap (subseq), and constituent overlap (const).
    Intensity of the \colorbox{cyan!50}{Blue} and \colorbox{red!50}{red} highlights corresponds to \emph{absolute}
    increase and decrease of scores with respect to MLE.
    \ours significantly improves results on challenging \nent examples without hurting performance on \ent,
    whereas \rmv improves scores on \nent at the cost of performance on \ent. 
}
\label{tab:hans}
\end{table}

\subsection{Word Overlap Bias}
\label{sec:hans}
We evaluate our method on word overlap bias in NLI. 
\citet{mccoy2019hans} show that models trained on MNLI largely rely on
word overlap between the premise and the hypothesis to make entailment predictions.
They created a challenge dataset (HANS) where premises may not entail high word-overlapping hypotheses.
Specifically, a model biased by word overlap would fail on three types of non-entailment examples:
(1) Lexical overlap,
        \eg \nl{The doctor visited the lawyer.} $\nRightarrow$ \nl{The lawyer visited the doctor.}.
(2) Subsequence,
        \eg \nl{The senator near the lawyer danced.} $\nRightarrow$ \nl{The lawyer danced.}.
(3) Constituent,
        \eg \nl{The lawyers resigned, or the artist slept.} $\nRightarrow$ \nl{The artist slept.}.

We evaluate both biased and debiased models on the three subsets of HANS and show F1 scores for each class in \reftab{hans}.
As expected, models trained by MLE almost always predict entailment (\ent), and thus performs poorly for the non-entailment class (\nent).
\ours improves performance on \nent in all cases with little degradation on \ent.
In contrast, \rmv improves performance on \nent at the cost of significant degradation on \ent.

Among all biased models, \hand produces the best debiasing results because it is designed to fit the word overlap bias, and indeed has zero recall on \nent when tested on HANS.
On the contrary, the improvement from \hypo is lower because it does not capture any word overlap bias.
Correspondingly, its performance drop on HANS is minimal compared to its in-distribution performance.
Among all debiased models, BERT has the best overall performance.
We hypothesize that pretraining on large data improves model robustness in addition to the debiasing effect from \ours.

\subsection{Stress Tests}
\begin{table}[ht]
    \centering
\footnotesize{
\setlength{\tabcolsep}{4pt}
\begin{tabular}{lrrrrrr}
\toprule
\multirow{2}{*}{method} & \multicolumn{3}{c}{Negation} & \multicolumn{3}{c}{Overlap} \\
                      \cmidrule(lr){2-4}                      \cmidrule(lr){5-7}
                      & \ent        & \con       & \neu       & \ent       & \con       & \neu \\
\midrule
\hypo                 & \bt{0.0}{41.2}  & \bt{0.0}{52.4}  & \bt{0.0}{50.5}  & \bt{0.0}{44.2}  & \bt{0.0}{52.8}  & \bt{0.0}{51.7}  \\
\cbow                 & \bt{0.0}{20.1}  & \bt{0.0}{48.2}  & \bt{0.0}{53.9}  & \bt{0.0}{49.7}  & \bt{0.0}{52.9}  & \bt{0.0}{55.6}  \\
\hand                 & \bt{0.0}{37.5}  & \bt{0.0}{45.0}  & \bt{0.0}{57.3}  & \bt{0.0}{56.7}  & \bt{0.0}{50.1}  & \bt{0.0}{57.8}  \\
\midrule
\multicolumn{7}{l}{\textbf{model: BERT}} \\
\midrule
MLE                   & \bt{0.0}{2.4}   & \bt{0.0}{81.1}  & \bt{0.0}{56.5}  & \bt{0.0}{19.2}  & \bt{0.0}{83.3}  & \bt{0.0}{59.4}  \\
\midrule
\ours-\hypo           & \bt{4.9}{7.3}   & \ws{0.3}{80.7}  & \ws{0.9}{55.6}  & \bt{8.3}{27.5}  & \ws{2.2}{81.1}  & \ws{0.4}{59.1}  \\
\ours-\cbow           & \bt{15.4}{17.9} & \bt{0.6}{81.7}  & \ws{1.0}{55.5}  & \ws{0.9}{18.3}  & \ws{3.3}{80.0}  & \ws{2.8}{56.6}  \\
\ours-\hand           & \bt{1.9}{4.3}   & \ws{0.5}{80.6}  & \ws{1.0}{55.5}  & \ws{4.1}{15.0}  & \ws{1.4}{81.9}  & \ws{2.0}{57.4}  \\
\midrule
\multicolumn{7}{l}{\textbf{model: DA}} \\
\midrule
MLE                   & \bt{0.0}{17.4}  & \bt{0.0}{47.3}  & \bt{0.0}{55.3}  & \bt{0.0}{46.7}  & \bt{0.0}{60.5}  & \bt{0.0}{57.8}  \\
\midrule
\ours-\hypo           & \ws{5.6}{11.8}  & \ws{0.3}{47.0}  & \ws{3.6}{51.8}  & \ws{5.1}{41.6}  & \ws{1.0}{59.4}  & \ws{2.2}{55.6}  \\
\ours-\cbow           & \bt{11.0}{28.4} & \ws{25.9}{21.4} & \ws{15.8}{39.5} & \ws{11.5}{35.2} & \ws{18.8}{41.7} & \ws{13.9}{43.8} \\
\ours-\hand           & \bt{7.3}{24.7}  & \ws{5.3}{42.0}  & \ws{9.0}{46.4}  & \ws{4.6}{42.2}  & \ws{4.5}{56.0}  & \ws{7.8}{49.9}  \\
\midrule
\multicolumn{7}{l}{\textbf{model: ESIM}} \\
\midrule
MLE                   & \bt{0.0}{12.0}  & \bt{0.0}{72.7}  & \bt{0.0}{54.6}  & \bt{0.0}{27.6}  & \bt{0.0}{76.4}  & \bt{0.0}{57.5}  \\
\midrule
\ours-\hypo           & \bt{10.8}{22.8} & \ws{5.0}{67.7}  & \ws{0.6}{54.0}  & \bt{9.9}{37.5}  & \ws{3.2}{73.2}  & \ws{0.8}{56.7}  \\
\ours-\cbow           & \bt{20.7}{32.7} & \ws{10.4}{62.3} & \ws{7.7}{46.9}  & \bt{2.8}{30.4}  & \ws{10.8}{65.6} & \ws{7.7}{49.8}  \\
\ours-\hand           & \bt{3.8}{15.8}  & \ws{8.1}{64.6}  & \ws{2.8}{51.8}  & \bt{11.6}{39.2} & \ws{5.7}{70.7}  & \ws{3.6}{53.9}  \\
\bottomrule
\end{tabular}
}
\caption{
F1 scores of each class on \stress.
Intensity of the \colorbox{cyan!50}{Blue} and \colorbox{red!50}{red} highlights corresponds to \emph{absolute}
increase and decrease of scores with respect to MLE.
\ours improves results on \ent (that exhibits label shift) with some degradation on other classes for DA and ESIM.
}
\label{tab:stress}
\end{table}

In addition to the word overlap bias exploited by HANS,
there are other known biases such as negation words and sentence lengths.
\citet{naik2018stress} conduct a detailed error anlaysis on MNLI
and create six stress test sets (\stress) targeting at each type of error.
We focus on the word overlap and negation stress test sets,
which expose dataset bias as opposed to model weakness according to \citet{liu2019inoculation}.
A model biased by word overlap rate and negation words are expected to have low accuracy on the entailment class on challenge data.
The complete results are shown in \refapp{stress}.

In \reftab{stress}, we show the F1 scores of each class for all models on \stress.\footnote{
Since results of \rmv are similar to those in \reftab{hans},
we put them in \refapp{stress}.}
Compared to results on HANS, \stress sees lower overall improvement from debiasing.
One reason is that \stress decreases word overlap \emph{rate} and injects negation words
by appending distractor phrases, i.e. \nl{true is true} and \nl{false is not true}.
While this introduces label shift on biased features, it also introduces covariate shift on the input.
For example, although \hand contains features designed to use word overlap rate (Jaccard similarity) and negation words,
its does not have big performance drop on the challenge data compared to its in-distribution performance,
showing that that distractor phrases may affect the model in other ways.

While all debiased models show improvement on \ent,
both DA and ESIM suffer from degradation on the other two classes,
especially when trained by \ours-\cbow.
We posit two reasons.
First, while \cbow is insufficient to represent complete sentence meaning,
it does encode a distribution of possible meanings.
Thus models debiased by \ours-\cbow might discard useful information.
Second, model capacity limits what is learned beyond a BOW representation.
DA shows the most degradation since it only uses local word interaction,
thus is essentially a BOW model.
In contrast, BERT has little degradation on in-distribution examples
regardless of the \sup classifier.


\section{Related Work and Discussion}
\label{sec:discussion}
\paragraph{Adversarial data collection.}
Aside from NLI, dataset bias has been exposed on benchmarks for other NLP tasks as well,
\eg paraphrase identification~\citep{zhang2019paws,zhang2019selection},
story close test~\cite{schwartz2017roc},
reading comprehension~\cite{kaushik2018much},
coreference resolution~\cite{zhao2018gender},
and visual question answering~\cite{agrawal2016analyzing}.
Most bias is resulted from artifacts in the data selection procedure and shortcuts taken by crowd workers.
To systematically minimize bias during data collection, adversarial filtering methods~\cite{sakaguchi2019winogrande,zellers2019hellaswag} have been proposed to
discard examples predicted well by a simple classifier.
This is similar to the \rmv baseline, except that we apply ``filtering'' at training time.
In general, our debiasing methods are complementary to adversarial data collection methods.

\paragraph{Debiased representation.}
Our work is closely related to the line of work on removing bias in data representations.
\citet{bolukbasi2016man,zhao2018neutral} learn gender-neutral word embeddings by forcing certain dimensions to be free of gender information.
Similarly, \citet{wang2019learning} construct a biased classifier and project its representation out of the model's representation.
For NLI, \citet{belinkov2019adversarial} use adversarial learning to remove hypothesis-related bias in the sentence representations.
However, for some NLP applications it may not be easy to separate biased features from useful semantic representations,
thus we correct the conditional distribution of the class label given these biased features instead of removing them from the input.
Concurrently, \citet{clark2019ensemble} take the same approach and further show its effectiveness on additional tasks including reading comparehension and visual question answering.

\paragraph{Distribution shift.}
Covariate shift~\cite{shimodaira2000improving, bendavid2006analysis} and label
shift~\cite{lipton2018detecting, zhang2013domain} are two well-studied settings under distribution shift,
which makes different assumptions on how $p(x, y)$ changes.
However, most works in these settings assume access to unlabeled data from the target distribution.
%
Our objective is more related to distributionally robust optimization~\cite{duchi2018learning, hu2018does},
which does not assume access to target data and optimizes the worst-case performance under \emph{unknown}, bounded distribution shift.
In contrast, we leverage prior knowledge on potential dataset bias.

\paragraph{Data augmentation.}
An effective way to tackle the challenge datasets is to train or finetune on similar
examples~\cite{mccoy2019hans, liu2019inoculation, jia2017adversarial},
which explicitly correct the training data distribution.
However, constructing challenge examples often rely on handcrafted rules that target a specific type of bias,
\eg swapping male and female entities~\cite{zhao2018gender,zhao2019gender},
synonym/antonym substitution~\cite{glockner2018breaking},
and syntactic rules~\cite{mccoy2019hans, ribeiro2018sears},
and may require human verification~\cite{zhang2019paws,jia2017adversarial}.
Data augmentation provides a way to encode our prior knowledge on the task,
\eg swapping genders does not affect coreference resolution result,
and syntactic transformations may affect sentence meanings.
Therefore, a related direction is to develop generic augmentation techniques
with linguistic priors~\cite{andreas2019compositional, karpukhin2019synthetic}.

\section{Conclusion}
Across all different dataset biases, the fundamental problem is that the majority training examples are not representative of the real-world data distribution (including the challenge data),
thus minimizing the average training loss no longer accurately describes our objective.
In this paper, we tackle the problem by adapting the learning objective to focus on examples that cannot be easily solved by biased features.
We show that our debiasing method improves model performance on challenge data given \emph{known} dataset bias.
However, current improvements largely rely on task-specific prior knowledge,
thus an important next step is to develop more general methods that tackle different types of biases.

\section*{Acknowledgments}
Yanchao Ni worked on an earlier version of this project while he was at New York University.
We thank the GluonNLP community for their support on reproducing prior results.

\clearpage
\bibliographystyle{acl_natbib}
\bibliography{all}
\clearpage
\appendix
\begin{table*}[t]
    \footnotesize{
\begin{tabular}{llllllllllllll}
\toprule
                        &             & \multicolumn{3}{c}{Negation} & \multicolumn{3}{c}{Overlap} & \multicolumn{3}{c}{Antonym} & \multicolumn{3}{c}{Length} \\
\cmidrule(lr){3-5}  \cmidrule(lr){6-8} \cmidrule(lr){9-11} \cmidrule(lr){12-14}
                        &             & \ent        & \con       & \neu       & \ent       & \con       & \neu       & \ent       & \con       & \neu       & \ent       & \con     & \neu        \\
\midrule
\hypo & \multirow{3}{*}{MLE} & \bt{0.0}{41.2}  & \bt{0.0}{52.4}  & \bt{0.0}{50.5}  & \bt{0.0}{44.2}  & \bt{0.0}{52.8}  & \bt{0.0}{51.7}  & - & \bt{0.0}{40.5}  & - & \bt{0.0}{55.1} & \bt{0.0}{52.5} & \bt{0.0}{51.5} \\
\cbow   &        & \bt{0.0}{20.1}  & \bt{0.0}{48.2}  & \bt{0.0}{53.9}  & \bt{0.0}{49.7}  & \bt{0.0}{52.9}  & \bt{0.0}{55.6}  & - & \bt{0.0}{19.0}  & - & \bt{0.0}{21.9} & \bt{0.0}{55.5} & \bt{0.0}{49.4} \\
\hand &        & \bt{0.0}{37.5}  & \bt{0.0}{45.0}  & \bt{0.0}{57.3}  & \bt{0.0}{56.7}  & \bt{0.0}{50.1}  & \bt{0.0}{57.8}  & - & \bt{0.0}{28.2}  & - & \bt{0.0}{66.6} & \bt{0.0}{65.0} & \bt{0.0}{60.7} \\
\midrule
\multirow{4}{*}{BERT}   & MLE         & \bt{0.0}{2.4}   & \bt{0.0}{81.1}  & \bt{0.0}{56.5}  & \bt{0.0}{19.2}  & \bt{0.0}{83.3}  & \bt{0.0}{59.4}  & - & \bt{0.0}{66.0}  & - & \bt{0.0}{83.8} & \bt{0.0}{83.6} & \bt{0.0}{77.4} \\
\cmidrule{2-14}
                        & \ours-\hypo & \bt{4.9}{7.3}   & \ws{0.3}{80.7}  & \ws{0.9}{55.6}  & \bt{8.3}{27.5}  & \ws{2.2}{81.1}  & \ws{0.4}{59.1}  & - & \bt{9.4}{75.4}  & - & \bt{0.3}{84.1} & \ws{0.4}{83.2} & \ws{1.1}{76.3} \\
                        & \ours-\cbow & \bt{15.4}{17.9} & \bt{0.6}{81.7}  & \ws{1.0}{55.5}  & \ws{0.9}{18.3}  & \ws{3.3}{80.0}  & \ws{2.8}{56.6}  & - & \bt{9.3}{75.3}  & - & \ws{1.3}{82.4} & \ws{1.4}{82.3} & \ws{2.8}{74.6} \\
                        & \ours-\hand & \bt{1.9}{4.3}   & \ws{0.5}{80.6}  & \ws{1.0}{55.5}  & \ws{4.1}{15.0}  & \ws{1.4}{81.9}  & \ws{2.0}{57.4}  & - & \bt{10.0}{76.0} & - & \ws{2.3}{81.4} & \ws{1.1}{82.5} & \ws{2.5}{74.9}        \\
\cmidrule{2-14}
                        & \rmv-\hypo  & \bt{29.7}{32.1} & \ws{25.2}{55.9} & \ws{16.6}{39.9} & \bt{25.2}{44.4} & \ws{19.5}{63.8} & \ws{16.5}{43.0} & - & \bt{3.3}{69.3}  & - & \ws{10.9}{72.9} & \ws{12.7}{70.9} & \ws{25.0}{52.4} \\
                        & \rmv-\cbow  & \bt{31.2}{33.6} & \ws{19.5}{61.6} & \ws{13.8}{42.7} & \bt{10.2}{29.4} & \ws{18.1}{65.2} & \ws{14.8}{44.7} & - & \bt{19.1}{85.1} & - & \ws{14.1}{69.7} & \ws{22.8}{60.8} & \ws{21.7}{55.7} \\
                        & \rmv-\hand  & \bt{18.3}{20.7} & \ws{31.4}{49.7} & \ws{16.5}{40.0} & \bt{11.7}{30.9} & \ws{28.6}{54.7} & \ws{19.8}{39.6} & - & \bt{17.8}{83.8} & - & \ws{26.6}{57.2} & \ws{31.0}{52.6} & \ws{30.9}{46.5} \\
\midrule
\multirow{4}{*}{DA}     & MLE         & \bt{0.0}{17.4}  & \bt{0.0}{47.3}  & \bt{0.0}{55.3}  & \bt{0.0}{46.7}  & \bt{0.0}{60.5}  & \bt{0.0}{57.8}  & - & \bt{0.0}{59.8}  & - & \bt{0.0}{69.5} & \bt{0.0}{66.0} & \bt{0.0}{61.9} \\
\cmidrule{2-14}
                        & \ours-\hypo & \ws{5.6}{11.8}  & \ws{0.3}{47.0}  & \ws{3.6}{51.8}  & \ws{5.1}{41.6}  & \ws{1.0}{59.4}  & \ws{2.2}{55.6}  & - & \ws{2.4}{57.4}  & - & \ws{3.2}{66.4} & \ws{2.3}{63.7} & \ws{6.6}{55.3}        \\
                        & \ours-\cbow & \bt{11.0}{28.4} & \ws{25.9}{21.4} & \ws{15.8}{39.5} & \ws{11.5}{35.2} & \ws{18.8}{41.7} & \ws{13.9}{43.8} & - & \ws{2.0}{57.8}  & - & \ws{5.3}{64.3} & \ws{26.5}{39.4} & \ws{8.0}{53.9} \\
                        & \ours-\hand & \bt{7.3}{24.7}  & \ws{5.3}{42.0}  & \ws{9.0}{46.4}  & \ws{4.6}{42.2}  & \ws{4.5}{56.0}  & \ws{7.8}{49.9}  & - & \bt{12.6}{72.4} & - & \ws{21.2}{48.4} & \ws{8.3}{57.6} & \ws{10.7}{51.2} \\
\cmidrule{2-14}
                        & \rmv-\hypo  & \ws{2.5}{14.9}  & \ws{7.4}{39.9}  & \ws{10.0}{45.3} & \bt{5.3}{52.0}  & \ws{7.8}{52.6}  & \ws{11.7}{46.0} & - & \ws{3.2}{56.6}  & - & \ws{5.7}{63.9} & \ws{3.7}{62.2} & \ws{30.0}{32.0}  \\
                        & \rmv-\cbow  & \ws{13.6}{3.8}  & \ws{23.4}{23.9} & \ws{17.3}{38.0} & \ws{44.2}{2.6}  & \ws{43.3}{17.1} & \ws{16.2}{41.5} & - & \ws{6.7}{53.1}  & - & \ws{65.2}{4.4} & \ws{32.4}{33.5} & \ws{32.0}{29.9} \\
                        & \rmv-\hand  & \bt{14.2}{31.6} & \ws{20.9}{26.4} & \ws{19.1}{36.2} & \ws{6.1}{40.6}  & \ws{22.8}{37.6} & \ws{28.1}{29.6} & - & \ws{2.0}{57.8}  & - & \ws{29.6}{40.0} & \ws{38.4}{27.6} & \ws{28.8}{33.1} \\
\midrule
\multirow{4}{*}{ESIM}   & MLE         & \bt{0.0}{12.0}  & \bt{0.0}{72.7}  & \bt{0.0}{54.6}  & \bt{0.0}{27.6}  & \bt{0.0}{76.4}  & \bt{0.0}{57.5}  & - & \bt{0.0}{75.1}  & - & \bt{0.0}{77.6} & \bt{0.0}{76.8} & \bt{0.0}{68.8} \\
\cmidrule{2-14}
                        & \ours-\hypo & \bt{10.8}{22.8} & \ws{5.0}{67.7}  & \ws{0.6}{54.0}  & \bt{9.9}{37.5}  & \ws{3.2}{73.2}  & \ws{0.8}{56.7}  & - & \bt{0.4}{75.5}  & - & \ws{1.7}{75.9} & \ws{2.5}{74.3} & \ws{2.5}{66.3} \\
                        & \ours-\cbow & \bt{20.7}{32.7} & \ws{10.4}{62.3} & \ws{7.7}{46.9}  & \bt{2.8}{30.4}  & \ws{10.8}{65.6} & \ws{7.7}{49.8}  & - & \ws{8.1}{67.0}  & - & \ws{9.1}{68.5} & \ws{16.6}{60.2} & \ws{8.7}{60.1} \\
                        & \ours-\hand & \bt{3.8}{15.8}  & \ws{8.1}{64.6}  & \ws{2.8}{51.8}  & \bt{11.6}{39.2} & \ws{5.7}{70.7}  & \ws{3.6}{53.9}  & - & \ws{0.4}{74.7}  & - & \ws{8.8}{68.8} & \ws{6.0}{70.9} & \ws{7.2}{61.6}    \\
\cmidrule{2-14}
                        & \rmv-\hypo  & \bt{17.6}{29.6} & \ws{18.3}{54.4} & \ws{9.3}{45.3}  & \bt{19.8}{47.3} & \ws{12.8}{63.6} & \ws{11.5}{46.1} & - & \ws{14.7}{60.4} & - & \ws{7.0}{70.6} & \ws{8.6}{68.2} & \ws{37.7}{31.1}  \\
                        & \rmv-\cbow  & \bt{19.8}{31.8} & \ws{40.7}{32.0} & \ws{25.7}{28.9} & \ws{9.5}{18.1}  & \ws{43.2}{33.2} & \ws{24.8}{32.7} & - & \ws{6.8}{68.3}  & - & \ws{51.0}{26.6} & \ws{58.8}{18.0} & \ws{28.0}{40.8} \\
                        & \rmv-\hand  & \bt{14.0}{26.0} & \ws{37.5}{35.1} & \ws{13.9}{40.7} & \bt{1.6}{29.2}  & \ws{33.1}{43.3} & \ws{23.5}{34.0} & - & \ws{17.7}{57.4} & - & \ws{46.2}{31.4} & \ws{41.8}{35.0} & \ws{33.5}{35.3} \\
\bottomrule
\end{tabular}
}
\caption{Complete results on MNLI Stress Test.}
\label{tab:app-stress}
\end{table*}

\section{Results on MNLI Stress Test}
\label{sec:stress}
In \reftab{app-stress},
we show the complete results on MNLI Stress Test~\cite{naik2018stress}.
In addition to Overlap and Negation,
which is intended to test dataset bias,
we also include two tests that evaluate model performance on minority examples.
Our debiased models have some improvement on Antonym,
possibly as a by-product of focusing on challenge examples that cannot be solved by superficial cues.
However, \ours did not improve performance on Length.

\end{document}